\pdfoutput=1

\documentclass[11pt]{article}

\usepackage[]{EMNLP2022}
\usepackage{subfig,graphicx}

\usepackage{times}
\usepackage{latexsym}

\usepackage[T1]{fontenc}

\usepackage[utf8]{inputenc}

\usepackage{microtype}

\usepackage{inconsolata}

\title{Evaluation of large-scale synthetic data for Grammar Error Correction}

\author{Vanya Bannihatti Kumar \\
  \texttt{vanya.bk@bytedance.com}\\} 

\begin{document}
\maketitle
\begin{abstract}
Grammar Error Correction(GEC) mainly relies on the availability of high quality of large amount of synthetic parallel data of grammatically correct and erroneous sentence pairs. The quality of the synthetic data is evaluated on how well the GEC system performs when pre-trained using it. But this does not provide much insight into what are the necessary factors which define the quality of these data. So this work aims to introduce 3 metrics - reliability, diversity and distribution match to provide more insight into the quality of large-scale synthetic data generated for the GEC task, as well as automatically evaluate them.
Evaluating these three metrics automatically can also help in providing feedback to the data generation systems and thereby improve the quality of the synthetic data generated dynamically
\end{abstract}

\section{Introduction}
Grammar error correction systems focus on detecting possible grammar errors like spelling errors, punctuation errors, verb errors etc and correcting them. Recent success in GEC systems depends largely on the availability of a large amount(in the scale of several millions) of parallel data consisting of grammatically correct and incorrect sentences. Some examples of this include:
\begin{itemize}
    \item \citet{stahlberg-kumar-2021-synthetic} use error type tags from ERRANT\citep{bryant-etal-2017-automatic} to guide the synthetic data distribution so that the distribution of the synthetic data can be similar to that of the domain in which the GEC system is used. This synthetic dataset consists of 200 million parallel sentences and achieves near state-of-the-art results on the CoNLL-14\citep{ng-etal-2014-conll} dataset
\end{itemize} 
\begin{itemize}
    \item \citet{awasthi-etal-2019-parallel} introduces the rule-based synthetic data generation method for GEC by using a list of common replacement errors, common insertion errors, common deletion errors and verb errors, and applying these errors randomly to any part of the grammatically correct sentence. Using this method, a very large synthetic dataset can be easily created for better GEC results
\end{itemize} 
\begin{itemize}
    \item \citet{koyama-etal-2021-comparison} compares the results of using the synthetic data generated using different backtranslation models such as Transformers, CNN and LSTM models. Here, a pseudo data of nearly 9 million samples are created using each of the above models and the quality of the synthetic data is compared by how well the GEC model trained using these synthetic datasets, perform on test sets like BEA-test\citep{bryant-etal-2019-bea} etc.
\end{itemize} 
As mentioned previously, all these synthetic datasets are evaluated based on how the GEC models perform when trained using them. However, this does not capture essential metrics which can be used to improve the synthetic data generation. This work aims to introduce these metrics in terms of :
\begin{itemize}
    \item Reliability - This metric can help evaluate which part of the dataset truly resembles humans in terms of creating grammatical errors in a particular sentence
\end{itemize}
\begin{itemize}
    \item Diversity - Many synthetic data generation techniques like \citet{stahlberg-kumar-2021-synthetic} and \citet{zhou-etal-2020-improving-grammatical} show that diverse synthetic data can help improve the GEC systems. But these works do not accurately measure the diversity of the datasets. So, this work aims to evaluate the diversity of any synthetic dataset more accurately.
\end{itemize}
\begin{itemize}
    \item Distribution match - This metric is useful in determining which synthetic dataset to be used to train the GEC model for a particular domain. In this work, the experiments are done using datasets across different domains like the open domain and the novel domain to determine the accuracy of the metric
\end{itemize}

For the rest of the discussion, the experiments of the proposed metrics are conducted on the following synthetic datasets:
\begin{itemize}
    \item Tagged corruption model - \citet{stahlberg-kumar-2021-synthetic}
\end{itemize}
\begin{itemize}
    \item Backtranslation method - \citet{xie-etal-2018-noising}
\end{itemize}
\begin{itemize}
    \item Rule-based method -  \citet{awasthi-etal-2019-parallel}
\end{itemize}

The backtranslation method is applied on the novel domain to verify if the metrics still holds for different domains other than the open domain.

\section{Reliability}
\label{sec:reliable}
A large portion of the parallel data produced by synthetic methods like backtranslation, round-trip translation, tagged-corruption models, rule-based methods etc., consists of erroneous sentences which are not realistically generated by humans. For example, consider one of the sentences in the google c4 corpus:
\begin{quote}
    \textit{19. Develop a calendar of local and regional events in your locale and make your company visible in the areas most related to your company and your potential clients' interests . }
\end{quote}
It's corrupted sentence from the tagged-corruption model is as follows:
\begin{quote}
    \textit{19 develop development a calendar of local and regional events in your louce and make your company visible in the areas most relating to your company. (Environment + Agriculture) and potential interests .}
\end{quote}
The difference between the erroneous sentence and the original sentence is as follows:
\begin{quote}
    \textit{[19. Develop -> 19 develop development] a calendar of local and regional events in your [locale -> louce] and make your company visible in the areas most [related -> relating] to your company [and your potential clients' -> (Environment + Agriculture) and potential] interests . }
\end{quote}
From this example, it is quite clear that except [related->relating] and [locale->louce] edits, all the other edits are impractical and are most likely not made by humans. Since these examples do not mimic the annotated parallel data of erroneous sentences and the grammatically correct sentences, training a model with such a data do not give any improvements. 
\newline
To this end, this work aims to formalise a metric which measures the number of reliable examples in the synthetic dataset automatically. 
\newline

\begin{table}
\centering
\begin{tabular}{c|l|c}
\hline
\textbf{Error type} & \textbf{Original}  & \textbf{Reliable} \\
\hline
OTHER&	23.45&	23.55\\
ADV&	1.79&	1.79\\
PREP&	9.16&	9.17\\
ORTH&	6.71&	6.74\\
NOUN&	16.11&	16.06\\
MORPH&	2.74&	2.74\\
DET&	8.04&	8\\
PRON&	1.72&	1.71\\
VERB:SVA&	1.28&	1.27\\
PART&	0.7&	0.69\\
VERB&	6.04&	6.02\\
VERB:TENSE&	3.62&	3.56\\
VERB:FORM&	2.4&	2.4\\
SPELL&	7.72&	7.77\\
CONJ&	0.9&	0.93\\
ADJ&	2.07&	2.07\\
WO&	0.81&	0.81\\
PUNCT&	1.63&	1.64\\
NOUN:NUM&	2.77&	2.76\\
ADJ:FORM&	0.08&	0.09\\
CONTR&	0.03&	0.03\\
NOUN:POS&	0.04&	0.04\\
NOUN:INFL&	0.09&	0.09\\
\hline
\end{tabular}
\caption{Distribution from tagged corruption model using ERRANT}
\label{tab:tag-corrupt}
\end{table}

To measure the  reliability metric of any synthetic data, the pre-trained binary classification BERT model from huggingface\cite{huggingface} was fine-tuned using the dataset created as follows:
\begin{itemize}
    \item The unreliable sentence pairs were created using rule-based method where the following rules were applied:
    \begin{itemize}
        \item The verb error was introduced from the verb error list by replacing a verb with the verb errors of another completely different and random verb. For example, instead of replacing 'abandon' with 'abandoning'(which creates a more realistic error), 'abandon' was replaced with 'associating', to create an improbable error sentence from the original sentence.
        \item Similarly for replacement error, the replacement list was used where instead of replacing the word in the replacement list with its commonly confused word, it was replaced with a commonly confused counterpart of a completely different and random word in the replacement list. For example, instead of replacing 'equipment' with 'equipmet', 'equipment' was replaced with 'Therefofe'(commonly confused word of 'Therefore'), to create an impractical and unreliable sentence pair.
        \item For deletion error, a word from the commonly inserted words list was deleted randomly. And for insertion error, a word from the commonly deleted words list was inserted randomly.
        \newline
        (The verb error, replacement, insertion and deletion error lists were obtained from \citet{awasthi-etal-2019-parallel}
    \end{itemize}
    \item The reliable sentence pairs were obtained from human annotated GEC corpus such as Lang8, NUCLE, W\&I+LOCNESS and FCE datasets which are publicly available.
\end{itemize}
The unreliable and the reliable sentence pairs were used in the ratio of 1:1 to create a dataset of 500k sentence pairs, which was used to fine-tune the pre-trained binary classification BERT model. 
This fine-tuned BERT classification model is used to get the reliability metric for a sample of 100k sentence pairs of few synthetic datasets as shown in Table \ref{tab:reliability-metric}
\begin{table}
\centering
\begin{tabular}{lc}
\hline
\textbf{Dataset} & \textbf{Reliability metric}\\
\hline
Tagged corruption model & 17.63\% \\
Backtranslation method & 26.22\% \\
Rule-based method & 9.61\% \\ \hline
\end{tabular}
\caption{Reliability metric for sample of few synthetic datasets}
\label{tab:reliability-metric}
\end{table}

The authenticity of the reliability metric was measured by further using these classified datasets, one part containing unreliable sentence pairs and the other part containing reliable sentence pairs, to fine-tune a GEC model which was pre-trained using Lang-8(\citet{mizumoto-etal-2011-mining}, \citeauthor{tajiri-etal-2012-tense}), W\&I+LOCNESS(\citet{bryant-etal-2019-bea}, \citet{Granger2014TheCL}) and FCE\citep{yannakoudakis-etal-2011-new} datasets, and  then verifying the F0.5 metrics on GEC test dataset as shown in the Tables \ref{tab:reliability-authenticity1} and \ref{tab:reliability-authenticity2}. The reliable and the unreliable datasets used for fine-tuning were both made of 50k sentence pairs. 

Table \ref{tab:reliability-authenticity1} shows the results of fine-tuning reliable and unreliable datasets of synthetic data generated by tagged corruption model and rule-based methods on CoNLL-14 test set. While Table \ref{tab:reliability-authenticity2} shows the results of fine-tuning reliable and unreliable datasets of synthetic data generated by the backtranslation method in the novel domain on a test set made of 8k sentence pairs in the novel domain. From these tables we can see that for the same size of dataset used for fine-tuning and with the same origin i.e, generated using the same synthetic means, there is a vast difference in how the GEC models fine-tuned on these datasets perform as measured by the F0.5 metric. There is a difference of at least 1.7\% in F0.5 between the GEC model trained on reliable dataset as compared to the unreliable one. This shows that the classification model is fairly accurate in distinguishing the reliable synthetic data from the unreliable ones. This is true not only in the open domain as shown in the Table \ref{tab:reliability-authenticity1} but across other domains like the novel domain as shown in the Table \ref{tab:reliability-authenticity2}.

\begin{table}
\centering
\begin{tabular}{l|c|c}
\hline
\textbf{Dataset} & \textbf{Reliable}  & \textbf{Unreliable} \\
\hline
& F0.5 & F0.5 \\
\hline
Tagged corruption  & 40.64\% & 36.6\% \\
Rule-based  & 48.35\% & 46.67\% \\ \hline
\end{tabular}
\caption{Performance of reliable and unreliable datasets from tagged corruption model and rule-based methods on CoNLL-14}
\label{tab:reliability-authenticity1}
\end{table}

\begin{table}
\centering
\begin{tabular}{lcc}
\hline
\textbf{Dataset} & \textbf{Reliable}  & \textbf{Unreliable} \\
\hline
& F0.5 & F0.5 \\
\hline
Backtranslation  & 15.10\% & 10.25\% \\
\end{tabular}
\caption{Performance of reliable and unreliable datasets from backtranslation model on test data in novel domain}
\label{tab:reliability-authenticity2}
\end{table}

\begin{table}
\centering
\begin{tabular}{l|c|c}
\hline
\textbf{Error type} & \textbf{Original}  & \textbf{Reliable} \\
\hline
OTHER&	18.17&	23.67\\
ADV&	0.5&	0.2\\
PREP&	4.93&	2.53\\
ORTH&	14.87&	6.69\\
NOUN&	1.47&	0.84\\
MORPH&	3.07&	0.89\\
DET	&7.03&	2.19\\
PRON&	2.29&	0.99\\
VERB:SVA&	0.14&	0.16\\
PART&	0.13&	0.07\\
VERB&	3.09&	1.51\\
VERB:TENSE&	5.62&	2.15\\
VERB:FORM&	3.4&	1.06\\
SPELL&	0.64&	1.15\\
CONJ&	1.09&	0.96\\
ADJ&	0.96&	0.3\\
WO&	0.49&	0.2\\
PUNCT&	29.92&	52.48\\
NOUN:NUM&	1.6&	0.53\\
ADJ:FORM&	0.43&	0.14\\
CONTR&	0.05&	0.18\\
NOUN:POS&	0.02&	1.12\\
NOUN:INFL&	0.01&	0\\
\hline
\end{tabular}
\caption{Distribution from backtranslation model} 
\label{tab:backtrans}
\end{table}

\section{Diversity}
Diversity is another key factor which determines the quality of the synthetic data. The ERRANT tool is used to measure the error types of the synthetic datasets in the English language. Previous works like \citet{zhou-etal-2020-improving-grammatical} interpret that a dataset has more diversity in error types if the number of 'OTHER' or 'Unknown' error sentence pairs as annotated by ERRANT is more. This work aims to figure out how the percentage of these error types changes with dataset containing both reliable and unreliable data v/s when it contains only reliable sentence pairs. Since the authenticity of the reliability metric was established in the previous section \ref{sec:reliable}, the measure of diversity distribution with only reliable sentence pairs is far more accurate for any synthetic dataset, than that with both reliable and unreliable sentence pairs. Table \ref{tab:tag-corrupt} shows that the diversity of the original dataset containing both reliable and unreliable sentence pairs, obtained using tagged corruption model and that of the filtered dataset containing only reliable sentence pairs remains almost constant. But it changes drastically for the synthetic dataset obtained using backtranslation method as shown in Table \ref{tab:backtrans}(distribution annotated using ERRANT). This shows that the diversity distribution for different synthetic datasets sometimes vary when the unreliable sentence pairs are removed from the dataset.

\begin{figure}\centering
\subfloat[Tagged corruption model]{\label{a}\includegraphics[width=1\linewidth]{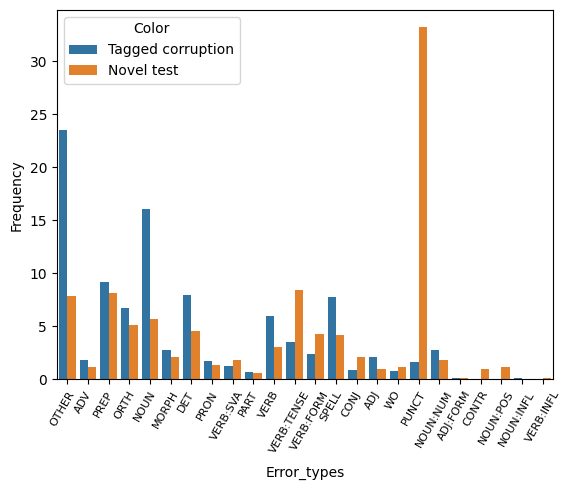}}\hfill
\subfloat[Backtranslation model]{\label{b}\includegraphics[width=1\linewidth]{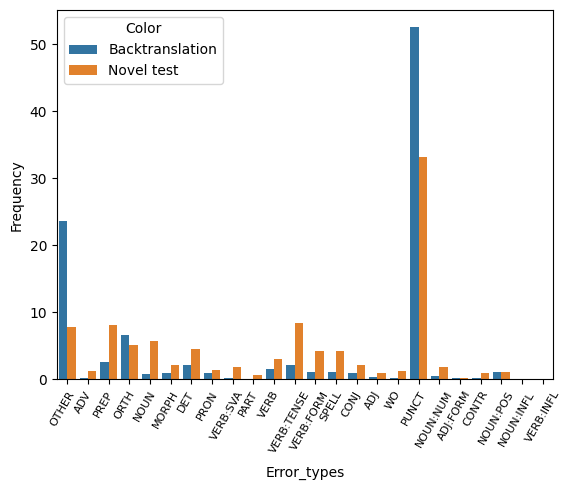}}\par 
\subfloat[Novel annotated dataset]{\label{c}\includegraphics[width=1\linewidth]{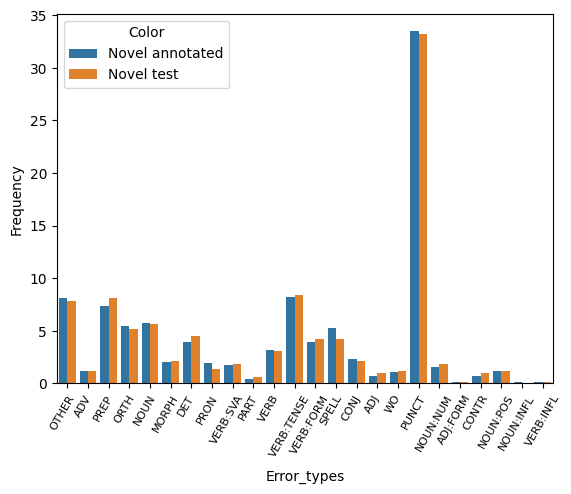}}
\caption{Comparing distributions of different datasets against the test set in novel domain}
\label{fig:comp}
\end{figure}

\section{Distribution match}
Distribution match of the synthetic dataset used to fine-tune the GEC model with that of the real-world test dataset is the most important factor which measures the quality of the synthetic set. If there is a large shift in the distribution between these two sets, the model trained using this synthetic dataset will perform worse as indicated in several previous works like \citet{stahlberg-kumar-2021-synthetic}. 
\newline
Experiments were conducted by fine-tuning the GEC model(pre-trained on Lang-8, WI+LOCNESS and FCE datasets) on different datasets having varying distributions. As is evident from Figure \ref{fig:comp}, the distribution of the synthetic datasets from tagged corruption model and backtranslation vary vastly from that of the novel test data distribution and hence the performance of the GEC model fine-tuned with these datasets is worse compared to the novel annotated dataset which has distribution very similar to that of the test set, as shown in Table \ref{tab:comp}.
\newline
Even though the synthetic datasets from tagged corruption model and backtranslation model were filtered to only include reliable sentences to match the quality of the annotated novel training dataset, ultimately since the distribution was much different compare to the test set, the performance was also much worse.

\begin{table}
\centering
\begin{tabular}{c|l|c|c}
\hline
\textbf{Dataset} & \textit{Precision}  & \textit{Recall} & \textit{F0.5} \\
\hline
Tagged Corruption & 12.86 & 21.74 & 14.01 \\
Backtranslation & 13.87 & 23.38 & 15.10 \\
Novel annotated & \textbf{50.66} & \textbf{33.98} & \textbf{46.13} \\
\hline
\end{tabular}
\caption{Results of GEC model fine-tuned on different datasets, evaluated against test set in novel domain} 
\label{tab:comp}
\end{table}

\section{Conclusion and Future Work}
This work formally introduces the three main parameters which are key to measuring the quality of a synthetic dataset. Several experiments were also conducted to measure the authenticity of these parameters and whether they are robust across domains. This ensures the universality of the parameters and can probably be extended to other similar tasks like machine translation. Since these parameters i.e, reliability, diversity and distribution match are measured automatically, they can be used as feedback to the system which generates synthetic data(much like in a Reinforcement Learning setting), thereby improving the overall quality of the synthetic dataset generated. With a very high quality of the synthetic data generated to either pre-train or fine-tune a GEC model, its performance would also improve drastically.

\section*{Limitations}
The limitations of this work is that since the reliability is measured automatically, large-scale annotated data for the classification task is required to obtain accurate classification results which may not be feasible. It would also require a lot of computing resources to evaluate on large scale synthetic datasets in the scale of millions.

\section*{Ethics Statement}
The datasets used for the experiments conducted in the paper are obtained from diverse sources like Lang-8, FCE, WI+LOCNESS etc, and different domains like novel, which includes data annotated by people whose first language is not English as well as from native English speakers, making it widely applicable in the society.


\bibliography{anthology,custom}
\bibliographystyle{acl_natbib}




\end{document}